\documentclass[letterpaper, 10 pt, conference]{ieeeconf}  

\IEEEoverridecommandlockouts                              

\usepackage{amsmath}
\usepackage{amssymb}
\usepackage{bm}
\usepackage{booktabs}
\usepackage{siunitx}
\usepackage{multirow}
\usepackage{multicol}
\usepackage{adjustbox}
\usepackage{subcaption}

\title{\LARGE \bf
Task-priority Intermediated Hierarchical Distributed Policies: Reinforcement Learning of Adaptive Multi-robot Cooperative Transport
}

\author{Yusei Naito$^{1}$, Tomohiko Jimbo$^{2}$, Tadashi Odashima$^{2}$, and Takamitsu Matsubara$^{1}$
\thanks{*This work was not supported by any organization}
\thanks{$^{1}$Y.~Naito and T.~Matsubara are with the Graduate School of Information Science, Nara Institute of Science and Technology (NAIST), Nara, Japan.}%
\thanks{$^{2}$T.~Jimbo and T.~Odashima are with the R-Frontier Division, Frontier Research Center, Toyota Motor Corporation, 1, Toyota-cho, Toyota, Aichi 471-8571, Japan}%
}

\begin{document}

\maketitle
\thispagestyle{empty}
\pagestyle{empty}

\begin{abstract}
Multi-robot cooperative transport is crucial in logistics, housekeeping, and disaster response. However, it poses significant challenges in environments where objects of various weights are mixed and the number of robots and objects varies. This paper presents Task-priority Intermediated Hierarchical Distributed Policies (TIHDP), a multi-agent Reinforcement Learning (RL) framework that addresses these challenges through a hierarchical policy structure. TIHDP consists of three layers: task allocation policy (higher layer), dynamic task priority (intermediate layer), and robot control policy (lower layer). Whereas the dynamic task priority layer can manipulate the priority of any object to be transported by receiving global object information and communicating with other robots, the task allocation and robot control policies are restricted by local observations/actions so that they are not affected by changes in the number of objects and robots. 
Through simulations and real-robot demonstrations, TIHDP shows promising adaptability and performance of the learned multi-robot cooperative transport, even in environments with varying numbers of robots and objects.
\end{abstract}

\section{INTRODUCTION}
Multi-robot cooperative transport, where robots collaborate to move objects, is gaining traction across logistics, household chores, and disaster response \cite{nath2020dist}\cite{tuci2018coop}. Take room tidying, for instance: lighter objects may be moved individually, while heavier ones require teamwork. However, predicting object weights beforehand is difficult due to the frequent replacement of objects, necessitating decisions on cooperation or individual handling. Hence, this study centers on multi-agent reinforcement learning for distributed control strategies, enabling multiple robots to cooperatively transport objects of various unknown weights in environments with wireless communication among robots.

Each robot repeatedly selects and transports an object in an environment with multiple objects. Therefore, the problem can be naturally divided into two parts: the task allocation problem for choosing which object to transport and the robot control problem for physical transport. 
An appropriate solution to solving these two problems simultaneously may be to learn a hierarchical policy: the higher policy for selecting objects and the lower policy for controlling the robot consistently. 

Applying Reinforcement Learning (RL) with such a hierarchical policy is challenging in environments where objects of various weights are mixed and the number of robots and objects varies, which is common in logistics, housekeeping, and disaster response applications. The higher policy inputs information about all entities and outputs object IDs for robot transportation. However, RL typically assumes fixed observation and action dimensions, complicating policy reuse with varying entity counts \cite{qie2019joint}\cite{niwa2022multi}. To address this, approaches with local observations and actions have been explored \cite{hsu2021scalable}\cite{shibata2023learning}; however, relying solely on local observations can hinder task allocation effectiveness, especially in large-scale environments. Additionally, local actions limit object selection, complicating adaptability versus performance balance.

\begin{figure}[t]
    \begin{subfigure}[b]{\linewidth}
        \includegraphics[width=\linewidth]{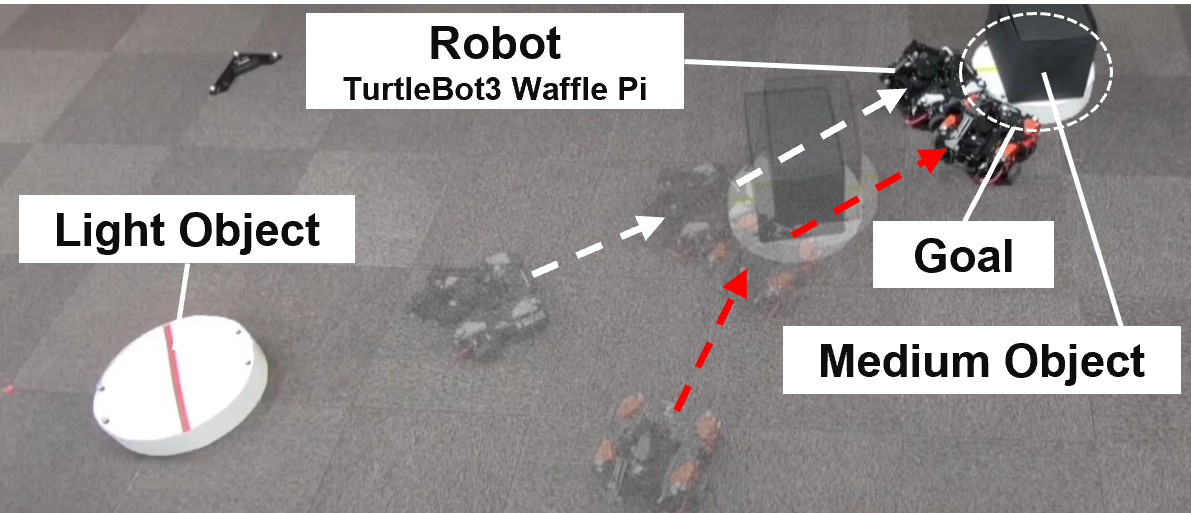}
        \caption{Cooperative action}
        \label{fig: real-coop}
    \end{subfigure}

    \vspace{2mm}

    \begin{subfigure}[b]{\linewidth}
        \includegraphics[width=\linewidth]{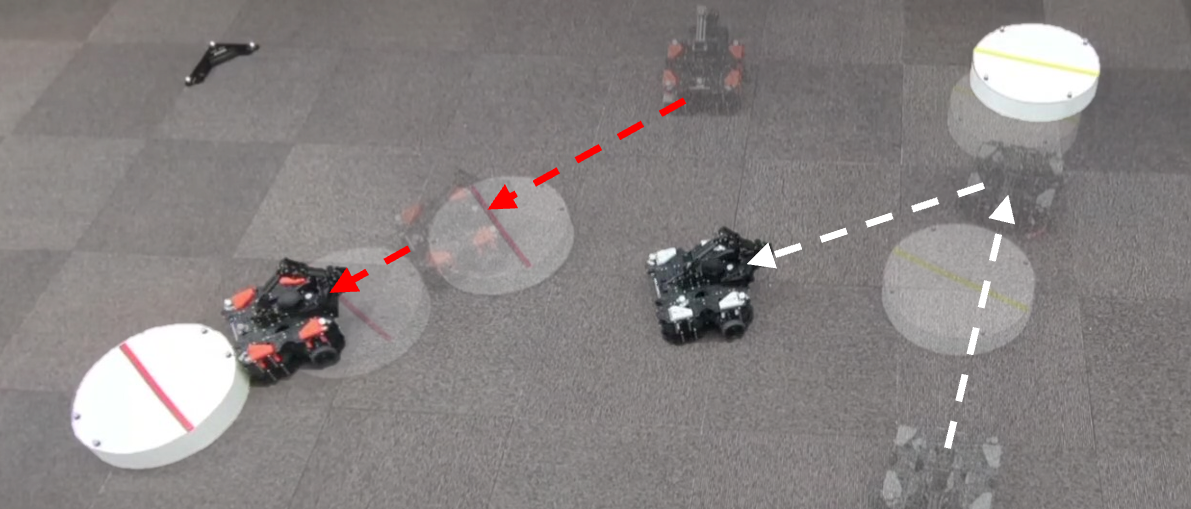}
        \caption{Independent action}
        \label{fig: real-indep}
    \end{subfigure}
    \caption{Multi-robot cooperative transport by our method. (a) Robots have high priority on one medium object and cooperate to transport it. (b) Each robot has a high priority for separate objects and transports them independently.}
    \label{fig: multi-robot-cooperative-transport}
\end{figure}

This paper introduces Task-priority Intermediated Hierarchical Distributed Policies (TIHDP), a multi-agent RL framework that addresses these challenges through a hierarchical policy structure. Each robot within this framework operates on a distributed policy structured into three layers: 1) task allocation policy (higher), 2) dynamic task priority (intermediate), and 3) robot control policy (lower). To maintain performance as the number of objects and robots in the environment changes, the task allocation and robot control policies are both restricted by local observations/actions so that they are not affected by changes in the number of objects and robots. The dynamic task priority layer, on the other hand, receives global object information and communicates with other robots to manipulate the priority of any object. The task allocation policy manages the priorities of objects in the neighborhood of the robots, while the task priority layer determines the highest priority object globally based on higher levels of influence and inter-robot communication. Finally, the robot control policy controls the behavior of the robot with respect to the target object. 
Through simulations and real-robot demonstrations (Fig.~\ref{fig: multi-robot-cooperative-transport}), TIHDP shows promising adaptability and performance in learning multi-robot cooperative transport tasks, even in environments with varying numbers of robots and objects.

\section{RELATED WORK}

\subsection{Combinatorial Optimization}

When employing deterministic methods in cooperative transportation, the task allocation can be formulated as a combinatorial optimization problem to minimize total travel distance under specified constraints on robot numbers. Techniques like the Hungarian algorithm \cite{Liu2011}  and integer linear programming \cite{sabattini2017} are utilized for this purpose. While these methods offer optimal solutions for minimizing travel distance, they often assume the availability of prior information regarding the required number of robots for each object. However, this assumption may not always hold. For instance, objects with frequent changes might necessitate manual data registration or estimation through cameras. Although cameras can provide object shape information, estimating required robot numbers based on weight poses challenges.

In contrast, the proposed method's strategy is crafted to determine subsequent actions relying on the robot's behavior and resulting changes in the position and speed of both the robot and object. This approach operates independently of prior knowledge about the necessary number of robots for transportation.




\subsection{Multi-agent Reinforcement Learning (MARL)}
Multi-agent Reinforcement Learning (MARL) has been explored for task allocation purposes \cite{qie2019joint}\cite{niwa2022multi}, utilizing Markov Decision Processes and Multi-agent Deep Deterministic Policy Gradient \cite{lowe2017multi}. However, these methods assume fixed numbers of robots and tasks, limiting their applicability beyond training scenarios. Addressing this, distributed policy models have been proposed, setting minimum thresholds for robots and tasks \cite{hsu2021scalable}. In multi-robot cooperative transport, frameworks employing dynamic task priorities and global communication have been effective \cite{shibata2023learning}, proving superior to using simplified rule-based algorithms for robot control. Cooperation in robot control, including force coordination and collision avoidance, is vital for real-world object transport. Existing research focuses on cooperative control and communication to position large objects \cite{shibata2021deep}.

Our proposed method employs MARL with a distributed policy model, limiting minimum robots and tasks. Distinguishing itself, it learns to handle multiple objects and specific robot controls for each transport, obviating the need for manual rule development.

\subsection{Hierarchical Reinforcement Learning}
Cooperative multi-object transport presents challenges due to the complexity of having each robot handle multiple objects. Learning such tasks, known for their long processes, is difficult \cite{mnih2015human}\cite{powell2007approximate}. Hierarchical reinforcement learning has been explored to address this, with goal-conditioned methods training lower-level policies to approach goal states set by higher-level policies and option framework-based methods training higher-level policies to switch between lower-level policies. These approaches have shown effectiveness in tasks like navigation \cite{nachum2018data}.

Our method deviates from prior work by introducing the dynamic task priority layer between the two policies, treating object IDs as targets in lower-level policies. Object IDs are represented discretely, inform lower-level observations and rewards, and enable the lower layer to focus on transporting individual objects as a reinforcement learning task.


\section{Problem Formulation}
The transport task targeted in this study consists of $N$ robots and $M$ objects, along with corresponding goals for these objects.
Here, both the robots and goals are assumed to be point mass models on a plane.

We first define the position and orientation, and the velocity and angular velocity of robot $i$ as $\bm{x}_i \in \mathbb{R}^3$ and $\bm{v}_i \in \mathbb{R}^3$, respectively.
Next, for object $l$, its position, goal position, velocity, and weight are defined as $\bm{z}_l \in \mathbb{R}^2$, $\bm{z}^*_l \in \mathbb{R}^2$, $\bm{w}_l \in \mathbb{R}^2$, and ${m}_l \in \mathbb{R}$, respectively.
Additionally, the radius of the goal is denoted as $D$.
Using these definitions, the transport of object $l$ is considered complete when $||\bm{z}_l - \bm{z}^*_l|| \leq D$.

Each robot is capable of observing the above-mentioned states for all robots.
Furthermore, they can observe all states of each object except for weight $m$.
Each robot’s policy outputs the control command $\bm{u}_i$ based on this information.

Additionally, the following assumptions are made:
\begin{quote}
  \begin{itemize}
    \item Robots know $M$ and $N$
    \item Robots can communicate globally with each other
  \end{itemize}
\end{quote}

\section{METHOD}

\begin{figure}[t]
    \centering
\includegraphics[width=1.0\linewidth]{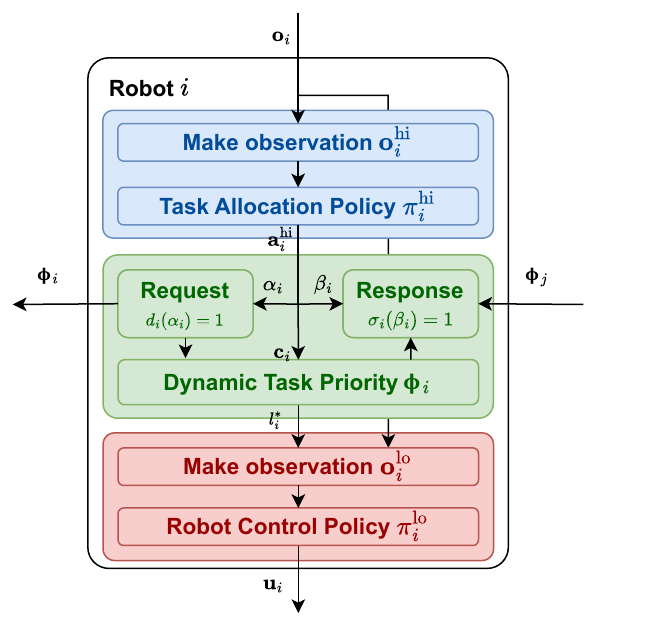}
    \caption{Task-priority Intermediated Hierarchical Distributed Policy. In robot $i$, the two policies of higher and lower use local observations $\bm{o}^{\text{hi}}_i$ and $\bm{o}^{\text{lo}}_i$. In the intermediate layer, task priorities $\bm{\Phi}_i$ are maintained while conducting global communication and the task priorities fluctuate when a request $\sigma_i(\beta_i)$ and response $d_i(\alpha_i)$ are established. The policy ultimately outputs the robot's control command $\bm{u}_i$.}

    \label{fig: process}
\end{figure}

This section presents our RL method with TIHDP for adaptive multi-robot cooperative transport, as shown in Fig.~\ref{fig: process}. 

\begin{figure*}[t]
    \centering
    \begin{subfigure}[b]{0.32\linewidth}
        \includegraphics[width=\linewidth]{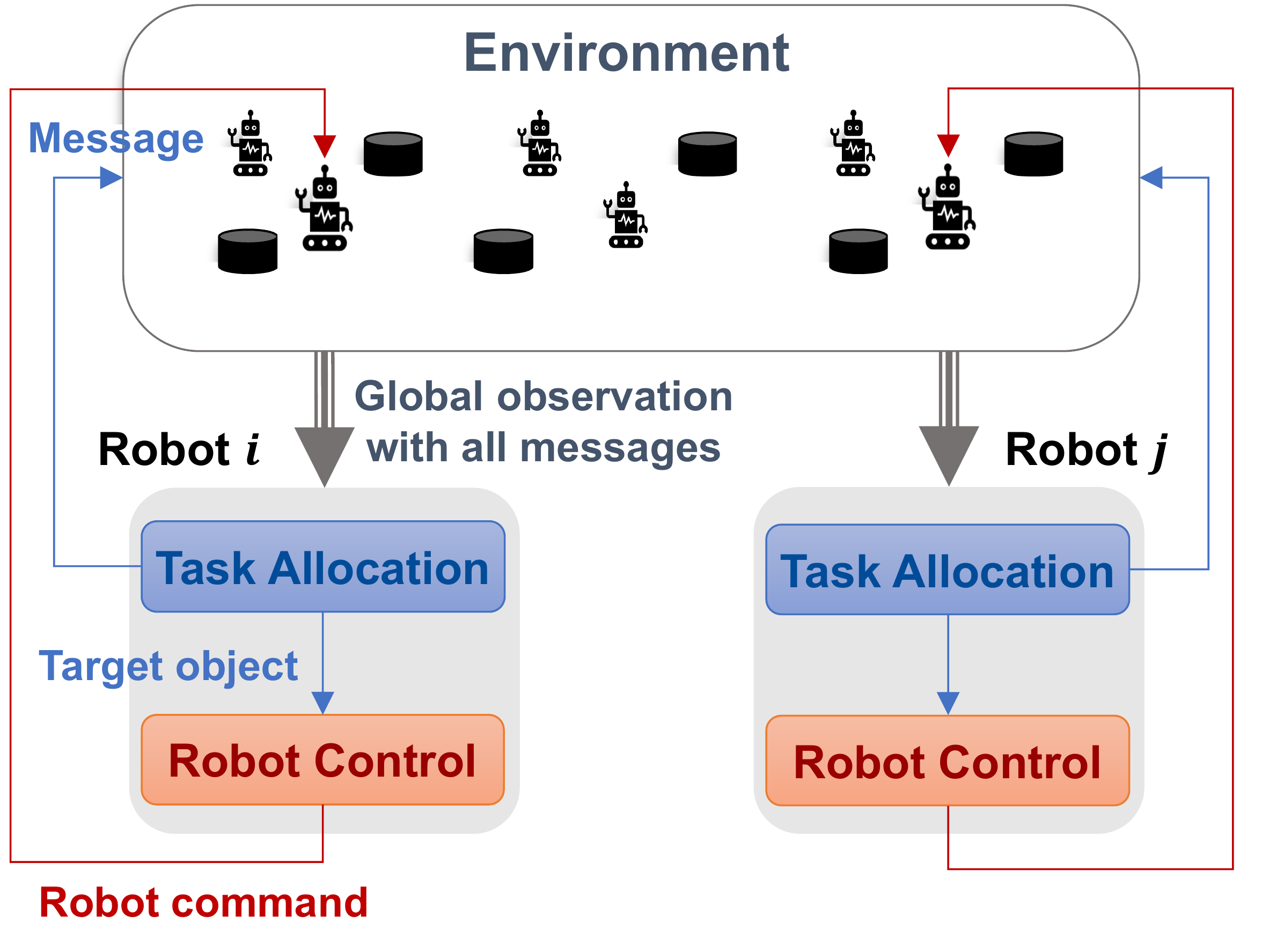}
        \vspace{2mm}
        \caption{Two-layered-global}
        \label{fig: 2lg}
    \end{subfigure}
    \begin{subfigure}[b]{0.32\linewidth}
        \includegraphics[width=\linewidth]{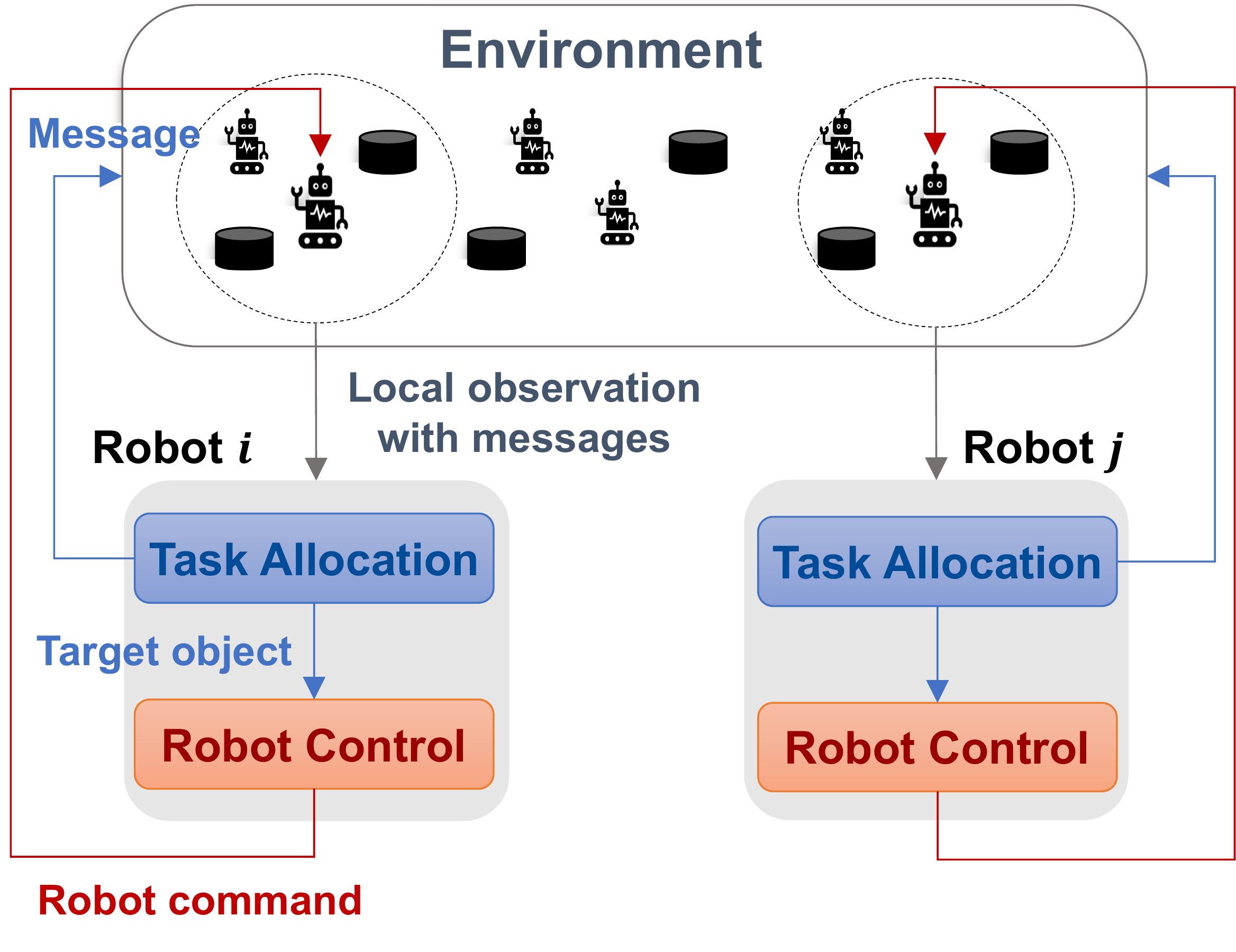}
        \vspace{1.4mm}
        \caption{Two-layered-local}
        \label{fig: 2ll}
    \end{subfigure}
    \begin{subfigure}[b]{0.32\linewidth}
        \includegraphics[width=\linewidth]{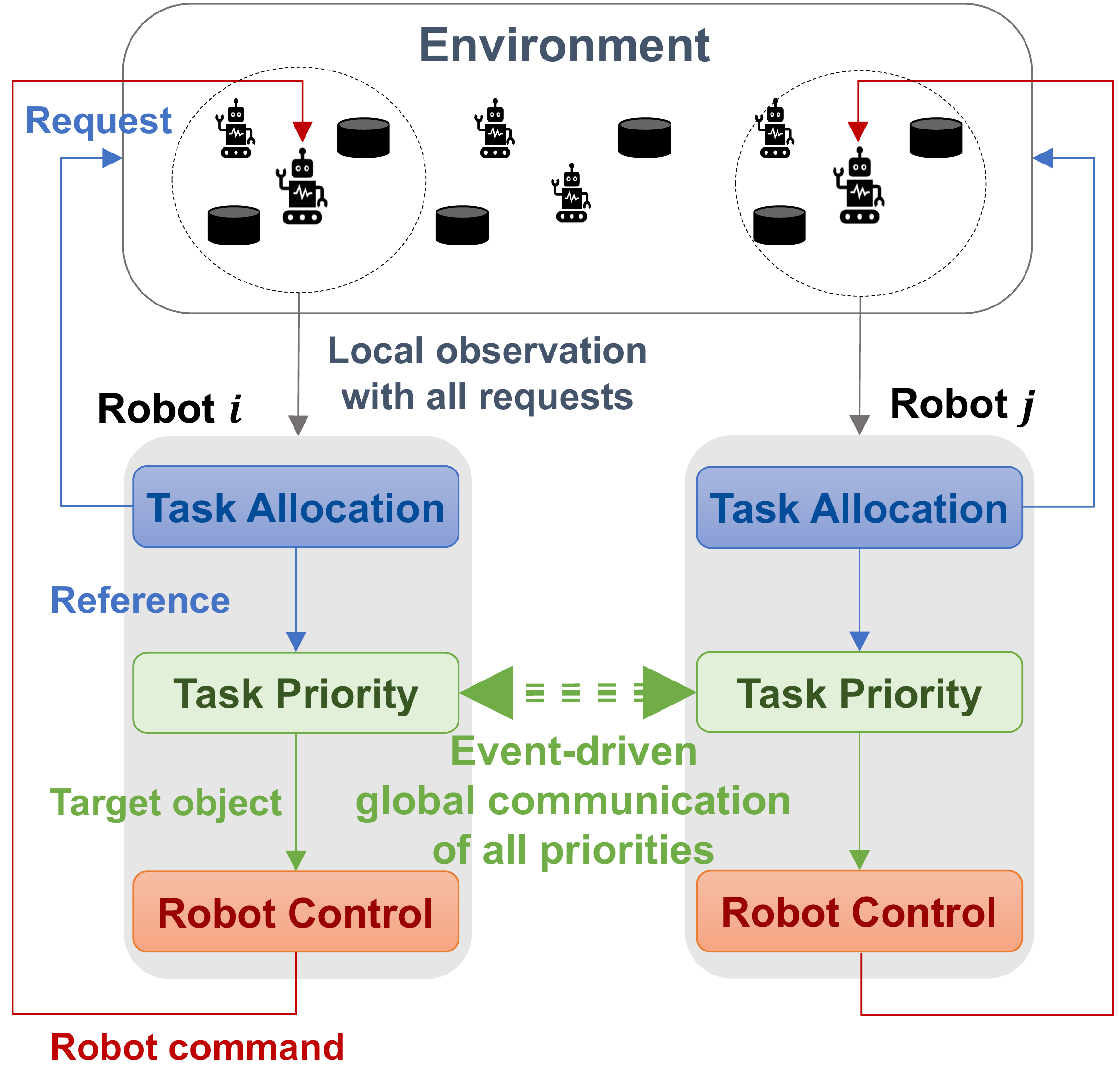}
        \caption{Ours}
        \label{fig: ours}
    \end{subfigure}
    \caption{Overview of frameworks. (a) Robots observe globally and select an object. (b) Robots observe locally and select an object from the vicinity. (c) Robots observe locally and update the task priority that possesses global memory.}
    \label{fig: proposal}
\end{figure*}

Robot $i$ initially receives an observation $\bm{o}_i$ from the environment. In the task allocation layer, $\bm{o}_i$ is condensed to a subset of nearby robots and objects, forming the higher-level observation $\bm{o}^{\text{hi}}_i$. The task allocation policy, $\pi^{\text{hi}}_i$, then generates a higher-level action $\bm{a}^{\text{hi}}_i$ based on this observation. Subsequently, in the dynamic task priority layer, the higher-level action is utilized to adjust the dynamic task priority $\bm{\phi}_i:=[\phi^1_i,...,\phi^M_i]\in\mathbb{R}^M$ through the robot's priority operation command $\bm{c}_i$.
Then, by comparing its response signal $\alpha_i$ with the request signal $\beta_j$ from robot $j$, the priority of robot $j$’s top-priority object $l_j^\ast$ in its list $\bm{\phi}_j$ is increased if a match occurs. Each robot focuses on transporting its highest-priority object $l_i^\ast$ from its list $\bm{\phi}_i$, using its observation $\bm{o}_i$ to derive a lower-level observation $\bm{o}^{\text{lo}}_i$ necessary for this task. Based on this observation, the lower-level policy $\pi^{\text{lo}}_i$ generates the robot command $\bm{a}^{\text{lo}}_i=\bm{u}_i$.


With such structured policies, both adaptability and performance to changes in the number of robots and objects can be achieved. More details are shown below.

\subsection{Distributed Partially Observable Markov Decision Process}
We utilize a distributed partially observable Markov decision process to apply MARL to multi-robot cooperative transportation. First, we define the state of the environment as $\bm{s}=[\bm{x},\bm{v},\bm{z},\bm{z}^{*},\bm{w},\bm{m}]$.
This includes information about all robots and objects. 

Next, for robot $i$, we define its observation and action as $\bm{o}_i$ and $\bm{a}_i$, respectively. The action $\bm{a}_i$ includes the robot's control command $\bm{u}_i$ and, if necessary, signals to other robots.
Each robot's distributed policy can observe information in the state $\bm{s}$, excluding the weight of the objects $\bm{m}$. However, to accommodate changes in the number of robots and objects, the number of input dimensions to the neural network must be constant. Therefore, the actual information about the robots and objects used is always selected in a fixed number according to each method.

Finally, we describe the reward design for evaluating actions that contribute to object transportation. The reward for object $l$, $r_l$, is given as follows:
\begin{eqnarray}
\label{eq:reward-rl}
  \begin{aligned}
    r_l= & \boldsymbol{w}_l\cdot\frac{\boldsymbol{z}^*_l-\boldsymbol{z}_l}{\|\boldsymbol{z}^*_l-\boldsymbol{z}_l\|}.
  \end{aligned}
\end{eqnarray}
As shown in Equation (\ref{eq:reward-rl}), the reward uses the component of the object's velocity in the direction of the goal. Therefore, a positive reward is obtained for actions that bring a robot closer to the object faster, while a negative reward (penalty) is obtained for actions that take it away from the object. To maximize the acquired reward, robots must always aim to acquire action policies that transport more objects to the goal faster.

\subsection{Task-priority Intermediated Hierarchical Distributed Policy Model}
\subsubsection{Task Allocation Layer}
Based on the method of Shibata et al. \cite{shibata2023learning}, each robot has a task allocation policy responsible for manipulating its dynamic task priorities and determining cooperative timings based on global communication.

As shown in Fig.~\ref{fig: ours}, the policy always observes a fixed number of nearby robots and objects, thus accommodating changes in the number of robots and objects. Specifically, as shown in Fig.~\ref{fig: process}, the task allocation layer of robot $i$ uses the positions $\bar{\bm{x}}_i$, velocities $\bar{\bm{v}}_i$ of the nearby $J$ robots, and the positions $\bar{\bm{z}}_i$, goal positions $\bar{\bm{z}}_i^\ast$, and velocities $\bar{\bm{w}}_i$ of the nearby $K$ untransported objects to create the higher-level observation $\boldsymbol{o}^{\text{hi}}_i=[\bm{u}_i, \bar{\bm{x}}_i, \bar{\bm{v}}_i, \bar{\bm{z}}_i, \bar{\bm{z}}_i^\ast, \bar{\bm{w}}_i]$.
Therefore, the distributed policy model for task allocation in this local observation is given as $\boldsymbol{a}^{\text{hi}}_i=[\boldsymbol{c_i},\alpha_i,\beta_i]=\pi^{\text{hi}}_i(\boldsymbol{o}^{\text{hi}}_i)$.

Here, $\boldsymbol{c}_i:=[c^1_i,...,c^K_i]\in \{-1, 1\}^K$ represents the manipulation of priorities $\phi^{l}_i$ for nearby objects, $\alpha_i\in \{0,1\}$ is the output for request signals, and $\beta_i\in \{0,1\}$ is the output for response signals.

\subsubsection{Dynamic Task Priority}
Based on the method of Shibata et al. \cite{shibata2023learning}, each robot has a dynamic task priority layer that updates the priorities for transporting each object.

As shown in Fig.~\ref{fig: process}, the dynamic task priority layer manipulates the priorities of each object's transportation based on higher-level actions and communication with other robots, identifying the object that should be transported at that moment.

First, we explain the operation of global communication. For robot $i$, request signals $d^l_i$ and response signals $\sigma_i$ for object $l$ are calculated using $\alpha_i$ and $\beta_i$ included in the higher-level actions as follows:

\begin{eqnarray}
  d^l_i(\alpha_i)=
  \begin{cases}
    1, & \text{if }\alpha_i =1 \text{ and } l=l^*_i \\
    0, & \text{otherwise }
  \end{cases},
\end{eqnarray}
\begin{eqnarray}
  \sigma_i(\beta_i)=
  \begin{cases}
    1, & \text{if }\beta_i =1 \\
    0, & \text{otherwise }
  \end{cases}.
\end{eqnarray}

These signals are always shared among all robots and affect the dynamic task priority when a request-response is established between robots.

Using the request signal $d_i^l$, response signal $\sigma_i$, and its own manipulation $\bm{c}_i$, the dynamic task priority is updated as follows:

\begin{align}
\label{eq:task-priority-update}
  \phi^{\prime}_{i}=
  \begin{cases}
    0, &\!\!\!\!\!\!\!\!\!\!\! \text{if } \|\boldsymbol{z}^*_l - \boldsymbol{z}_l\| < D \\
    (1-k_\phi)\phi^l_i+k_\phi \left( \bar{c}_i^l+\sigma_i\sum\limits_{j=1}^Nd^l_j\right), & \text{otherwise}
  \end{cases},
\end{align}

\begin{eqnarray}
\label{eq:task-priority-normalize}
  \phi^l_{i}=\frac{\phi^{l\prime}_i}{\sum_{l=1}^M\phi^{l\prime}_i}.
\end{eqnarray}
Here, $k_\phi>0$ is a constant that limits the amount of priority change per operation, and $\bar{\bm{c}}_i :=[\bar{c}^1_i,...,\bar{c}^M_i]\in \{-1,0,1 \}^M$ is a mapping of $\bm{c}_i$ in the order of object IDs, and values beyond the nearest $K$ neighbors are zero.

As shown in Equation (\ref{eq:task-priority-update}), the priority of an object that has been transported at that point is set to 0. The determination of whether the transportation is complete is constantly made, so an object that has been transported once will return as a transportation target if it moves away from the goal more than $D$ due to being touched by a robot.

Based on the manipulation $\boldsymbol{c}_i$ of the priorities of nearby objects, the priority of any object either increases or decreases. Furthermore, based on global communication, if the request signal $d^l_j$ for object $l$ from robot $j$ and the response signal $\sigma_i$ from robot $i$ satisfy $d^l_j=\sigma_i=1$, the priority $\phi_i^l$ of object $l$ held by robot $i$ increases. In other words, it is possible to raise the priority of objects not included in the higher-level observation $\bm{o}_i^{hi}$ in response to requests from other robots. Finally, as shown in Equation (\ref{eq:task-priority-normalize}), all task priorities are normalized so that their sum equals 1.

The object $l^*_i$ that robot $i$ should transport at that moment is selected as the one with the highest priority among all objects, as shown in the following equation:
\begin{eqnarray}
l^*_i = \underset{l}{\mathrm{argmax}} ~ \phi_i^l.
\end{eqnarray}

\subsubsection{Robot Control Layer}
Each robot has a robot control layer responsible for selecting the robot's control commands $\bm{u}_i$.

As shown in Fig.~\ref{fig: process}, the robot control layer for robot $i$ uses the currently required object $l^*_i$ for transportation to create the lower-level observation $\bm{o}_i^{\text{lo}}$. This includes information about the nearest object to avoid collisions. In other words, a single robot control policy is used for all objects. This allows the experience of trial and error in transportation to be shared across all objects, potentially improving sample efficiency during learning. For the robots, only the nearest information is observed. Therefore, the lower-level observation $\boldsymbol{o}^{\text{lo}}_i$ used by the robot control policy is given as $\boldsymbol{o}^{\text{lo}}_i=[\bm{u}_i, \bm{x}_{\text{nearest}_i}, \bm{v}_{\text{nearest}_i}, \bm{z}_{l^*_i}, \bm{z}^{*}_{l^*_i}, \bm{w}_{l^*_i}, \bm{z}_{\text{nearest}_i}, \bm{w}_{\text{nearest}_i}]$.
Thus, the distributed policy model for robot control in this observation $\boldsymbol{o}^{\text{lo}}_i$ is given as $\boldsymbol{a}^{\text{lo}}_i=\boldsymbol{u}_i=\pi^{\text{lo}}_i(\boldsymbol{o}^{\text{lo}}_i)$.

\subsection{Hierarchical Reward Design}
Reinforcement learning is conducted in both the task allocation layer and the robot control layer in a synchronized way; the task allocation layer uses a reward aimed at optimization for the entire team, while the robot control layer uses another reward focused on optimizing the transportation of a single object.

\subsubsection{Reward for Task Allocation Policy}
The reward $r^{\text{hi}}_i$ for the task allocation policy of robot $i$ is given as:
\begin{eqnarray}
  r^{\text{hi}}_i=\sum_{l=1}^{M}r_l.
\end{eqnarray}
This ensures that all robots receive the same reward at the same time. To maximize the acquired reward, cooperative behavior that considers other robots is necessary.

\subsubsection{Reward for Robot Control Policy}
The reward $r^{\text{lo}}_i$ for the robot control policy of robot $i$ is given as follows:
\begin{eqnarray}
  r^{\text{lo}}_i=r_{l^*_i}+\min(0, r_{\text{nearest}}) + e_i,
\end{eqnarray}
where,
\begin{eqnarray}
  e_i=
  \begin{cases}
  1, & \text{if } \bm{v}_i \cdot (\bm{z}_{l_i^\ast} - \bm{x}_i)>0 \text{ or } \|\bm{z}_{l_i^\ast} - \bm{x}_i\|<E \\
  0, & \text{otherwise}
  \end{cases}.
\end{eqnarray}
Here, $r_{l^i}$ is the reward for the object to be transported and $r_{\text{nearest}}$ is the reward for the nearest object. $E$ is a constant representing the distance at which an object is considered sufficiently close.

Thus, the robot control policy is trained with the goal of transporting the currently required object $l^*$ and avoiding collisions with the nearest object.

\subsection{Centralized Training with Decentralized Execution}
We used the Multi-agent Proximal Policy Optimization (MAPPO) in the learning of each policy \cite{yu2022surprising}. MAPPO is an extension of Proximal Policy Optimization (PPO) \cite{schulman2017proximal}, a reinforcement learning algorithm, to a centralized training with decentralized execution structure. The Actor network corresponding to the policy is distributed using local observations. The Critic network used during learning can uniquely use the observations of other robots as well, facilitating the acquisition of cooperative actions.




\section{EXPERIMENT}
\subsection{Experimental Overview}
Simulations were used to learn and validate the proposed method. There are two main objectives in the experiment.
The first is verifying whether the proposed method can learn the cooperative transport task's action policy. A physical simulation environment mimicking the actual characteristics of robots and objects was constructed for this.
The second is the performance evaluation of the policy after learning. The transport performance of the policy post-learning was compared with other learning methods. Additionally, performance evaluation was conducted in environments with a different number of robots and objects than those used during the learning phase.

\subsection{Simulation Environment}
NVIDIA's physics simulator, Isaac Sim, was used to construct a parallel simulation environment for cooperative transport, as shown in Fig.~\ref{fig:isaac-sim}. Furthermore, an environment for learning and executing the proposed method was implemented, based on Isaac Orbit \cite{mittal2023orbit}. Parallel simulation environments and their corresponding reinforcement learning libraries enable faster collection of experience data compared to traditional learning environments, as reported by \cite{handa2023dextreme}\cite{makoviychuk2021isaac}.

\subsection{Task Setting}
We used the TurtleBot3 Waffle Pi robot from ROBOTIS. TurtleBot3 has two drive wheels, one on each side. In this experiment, no actuators other than the drive wheels are used and the object is transported by pushing. The control command space is $\boldsymbol{u} = [u_{\text{move}}, u_{\text{turn}}]^T$, where $u_{\text{move}} \in \{\text{forward}, \text{backward}, \text{none}\}$ and $u_{\text{turn}} \in \{\text{right}, \text{left}, \text{none}\}$, allowing for combined movements of forward-backward motion and turning.

Objects are disc-shaped. All objects have the same shape but three different weights: light, medium, and heavy (which cannot be carried even with cooperation). The robot cannot distinguish the weight of an object by its shape, so it has to decide whether to cooperate or share the work only when it actually pushes the object.

\begin{figure}[t]
    \centering
    \begin{subfigure}[b]{0.49\linewidth}
        \includegraphics[width=\linewidth]{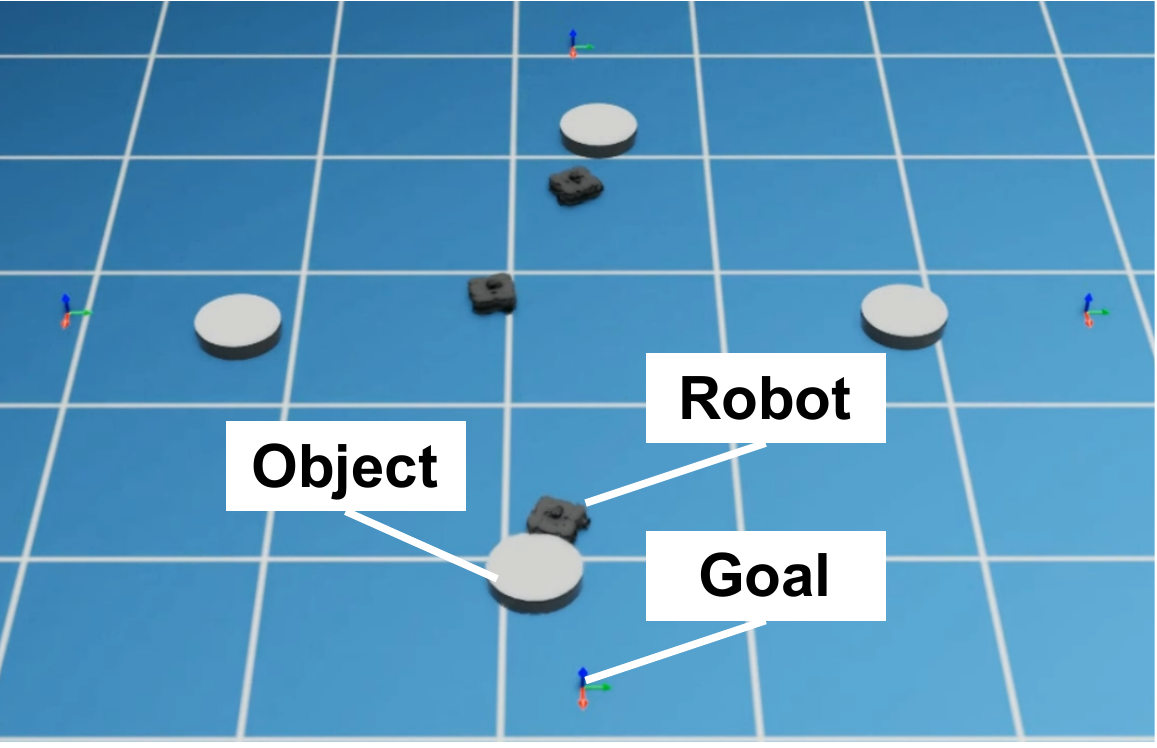}
        \caption{Rendered image}
        \label{fig:isaac-sim}
    \end{subfigure}
    \begin{subfigure}[b]{0.49\linewidth}
        \centering
        \includegraphics[width=0.6\linewidth]{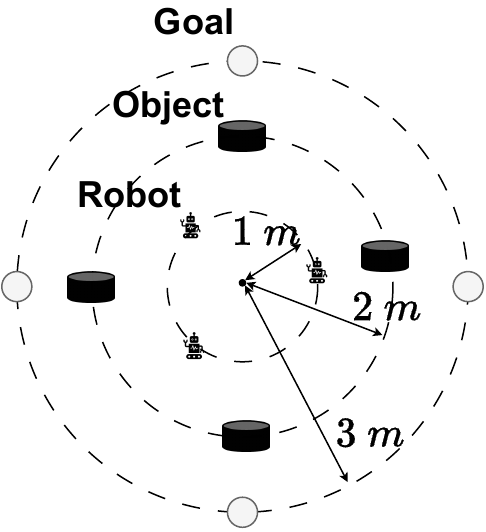}
        \caption{Initial placement}
        \label{fig:place}
    \end{subfigure}
    \caption{Simulation environment. Robots, objects, and goals are arranged in a circular pattern from the inside out.}
\end{figure}

The placement of robots, objects, and goals is shown in Fig.~\ref{fig:place}. In the initial state, the robots are positioned randomly within a \SI{0.5}{m} radius in the front, back, left, or right direction from a reference position on the circumference of a circle with a radius of \SI{1}{m}. The objects are placed randomly within a \SI{0.3}{m} radius in the front, back, left, or right direction from a reference position on the circumference of a circle with a radius of \SI{2}{m}. Each object has its own goal, which is positioned on the circumference of a circle with a radius of \SI{3}{m}.

\subsection{Evaluation Metrics}
To quantitatively evaluate the performance of the method, the following metrics are used:
\begin{itemize}
\item Completed Object Ratio (COR): Proportion of transportable objects that have been successfully transported.
\item Total Object Completion Ratio (TOCR): Proportion of all transportable objects that have been successfully transported.
\end{itemize}

\subsection{Comparison Methods}
The following two types, consisting of a task allocation layer and a robot control layer, were used for comparison:
\begin{itemize}
    \item Two-layered-global: with global observation and action  (Fig.~\ref{fig: 2lg})
    \item Two-layered-local: with local observation and action (Fig.~\ref{fig: 2ll})
\end{itemize}

\subsection{Results}
\subsubsection{Transport Performance}
\begin{figure}[t]
\centering
\includegraphics[width=0.8\linewidth]{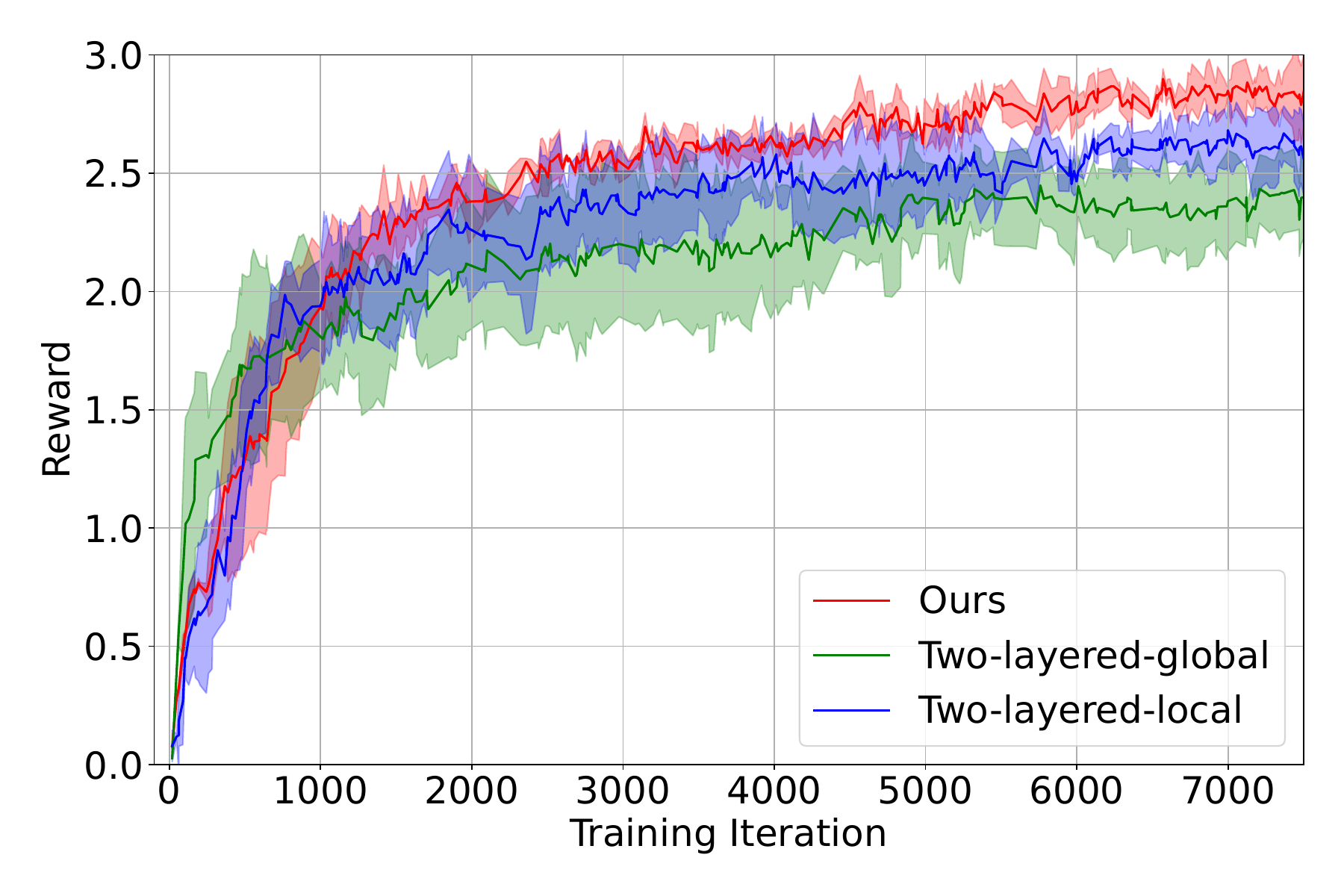}
\vspace{-3mm}
\caption{Cumulative rewards of evaluated methods}
\label{fig:reward-mappo-ours}
\end{figure}

The rewards for the proposed method and two comparison methods during training using the settings in Table~\ref{tab:learning-settings} and Table~\ref{tab:learning-policy-parameters} are shown in Fig.~\ref{fig:reward-mappo-ours}. Furthermore, the results of evaluating the completion rate of transport are presented in Table~\ref{tab:transported-rate}.
As shown in Table 3, our proposed method and Two-layered-local can be applied to any scenario different from the training scenarios, whereas Two-layered-global only applies to scenarios with the same number of robots and objects as in the training phase. Moreover, it was found that the proposed method has a higher transport performance compared to the two methods. The Two-layered-global method may have encountered difficulties in learning due to the size of the search space and action space. Moreover, it appears that the Two-layered-local method, lacking the capability to globally comprehend the environment, faced challenges in making appropriate decisions.

\begin{table}[t]
\centering
\caption{Simulation parameters for training}
\label{tab:learning-settings}
\begin{tabular}{lcwc{20truemm}}
\toprule
\multicolumn{2}{c}{\textbf{Parameter}} & \textbf{Value}\\
\midrule
Number of robots & $N$ & 3\\
Number of objects & $M$ & 4\\
Lightweight objects & & 2 \\
Medium-weight   objects & & 1\\
Heavyweight objects & & 1\\
Goal radius& $D$ & \SI{0.1}{m}\\
Episode length &  & 400 steps\\
\bottomrule
\end{tabular}
\end{table}

\begin{table}[t]
\centering
\caption{Policy parameters}
\label{tab:learning-policy-parameters}
\scalebox{0.8}{
\begin{tabular}{lcc}
\toprule
\multicolumn{2}{c}{\textbf{Parameter}} & \textbf{Value} \\
\midrule
Task Allocation Policy Hidden Layers&& [256, 128, 64] with LSTM \\
Robot Control Policy Hidden Layers&& [256, 128, 64] \\
Task Priority Time Constant& \(k_\phi\) & 0.1 \\
\bottomrule
\end{tabular}
}
\end{table}

\begin{table}[t]
\centering
\caption[Evaluation of transport performance]{Evaluation of transport performance where L represents lightweight objects, M medium-weight objects, and H heavyweight objects.}
\label{tab:transported-rate}
\scalebox{0.9}{
\begin{tabular}{wc{3truemm}wc{3truemm}wc{3truemm}ccwc{7truemm}wc{7truemm}wc{7truemm}wc{7truemm}}
\toprule
\multicolumn{3}{c}{\textbf{Objects}} & & & \multicolumn{2}{c}{\textbf{Two-layered}} & \multicolumn{2}{c}{\textbf{Ours}} \\ 
\textbf{\# L}        & \textbf{\# M}         & \textbf{\# H}        & \textbf{\# Robots} & \textbf{Metrics} & \textbf{global} & \textbf{local} & \textbf{w/o com} & \textbf{w/com} \\ \midrule
\multirow{2}{*}{2}   & \multirow{2}{*}{1}    & \multirow{2}{*}{1}   & \multirow{2}{*}{3} & COR              & 0.638                       & 0.737                      & 0.805    & \textbf{0.815}  \\
                     &                       &                      &                    & TOCR             & 0.180                       & 0.281                      & 0.430    & \textbf{0.500}  \\ \midrule
\multirow{2}{*}{3}   & \multirow{2}{*}{1}    & \multirow{2}{*}{1}   & \multirow{2}{*}{3} & COR              & -                           & 0.757                      & 0.791    & \textbf{0.805}  \\
                     &                       &                      &                    & TOCR             & -                           & 0.172                      & 0.335    & \textbf{0.367}  \\ \midrule
\multirow{2}{*}{3}   & \multirow{2}{*}{2}    & \multirow{2}{*}{1}   & \multirow{2}{*}{3} & COR              & -                           & 0.584                      & 0.634    & \textbf{0.641}  \\
                     &                       &                      &                    & TOCR             & -                           & 0.0234                     & 0.0313   & \textbf{0.0625} \\ \midrule
\multirow{2}{*}{3}   & \multirow{2}{*}{2}    & \multirow{2}{*}{1}   & \multirow{2}{*}{4} & COR              & -                           & 0.652                      & 0.686    & \textbf{0.705}  \\
                     &                       &                      &                    & TOCR             & -                           & 0.00781                    & 0.0938   & \textbf{0.109}  \\ \bottomrule
\end{tabular}
}
\end{table}

\subsubsection{Cooperative and Divided Actions}
\begin{figure*}[tbp]
    \centering
        \begin{subfigure}[b]{0.24\linewidth}
            \includegraphics[width=\linewidth]{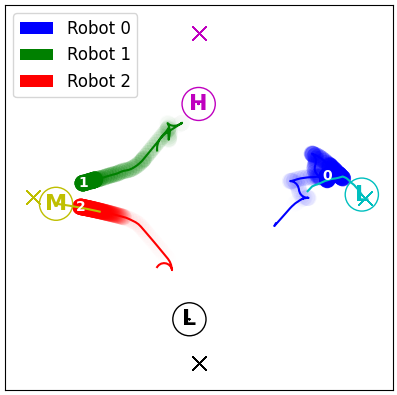}
            \caption{0 - \SI{10}{s}}
            \label{fig:trj-a}
        \end{subfigure}
        \begin{subfigure}[b]{0.24\linewidth}
            \includegraphics[width=\linewidth]{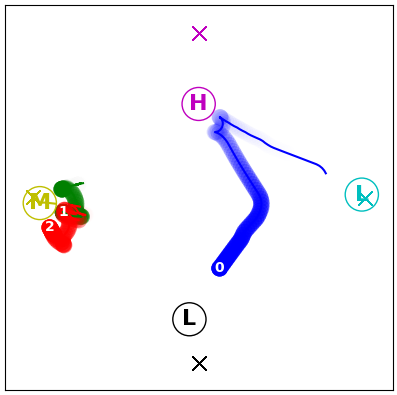}
            \caption{10 - \SI{20}{s}}
            \label{fig:trj-b}
        \end{subfigure}
        \begin{subfigure}[b]{0.24\linewidth}
            \includegraphics[width=\linewidth]{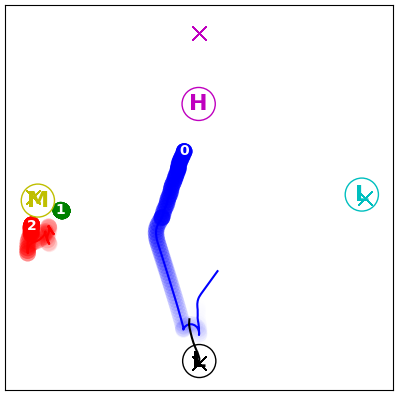}
            \caption{20 - \SI{30}{s}}
            \label{fig:trj-c}
        \end{subfigure}
        \begin{subfigure}[b]{0.24\linewidth}
            \includegraphics[width=\linewidth]{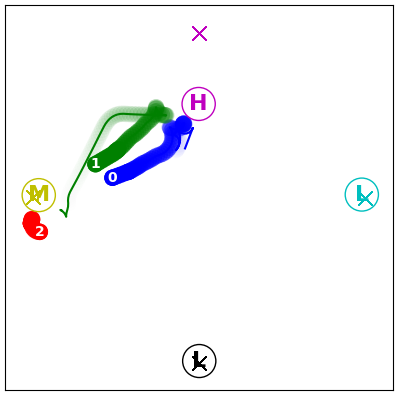}
            \caption{30 - \SI{40}{s}}
            \label{fig:trj-d}
        \end{subfigure}
        \begin{subfigure}[b]{\linewidth}
            \vspace{2mm}
            \includegraphics[width=\linewidth]{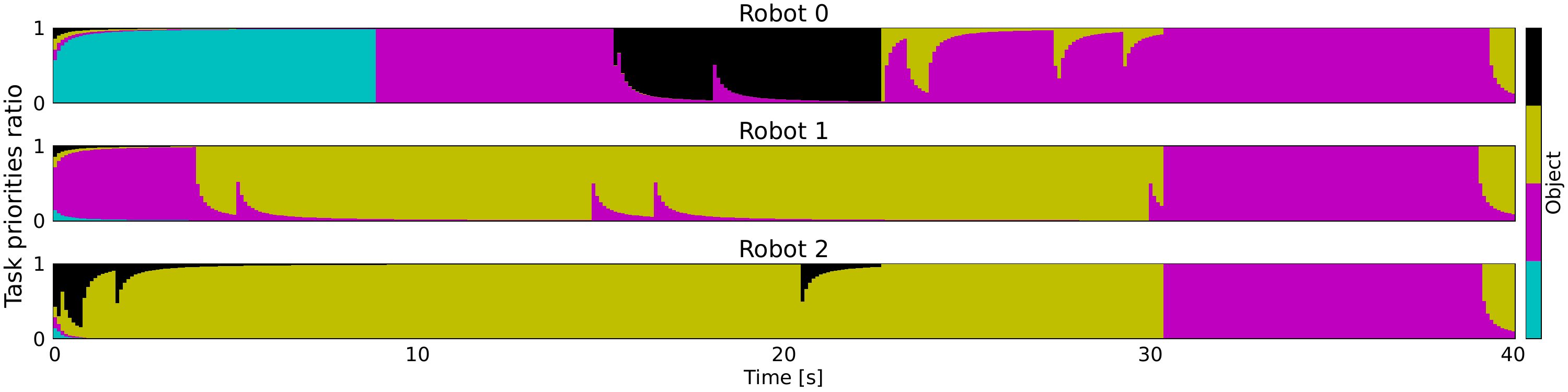}
            \caption{Task priorities}
            \label{fig:trj-priority}
        \end{subfigure}
    \caption{Trajectories of robots and objects. (a)-(d) The colored circles, cross marks, and letters indicate the objects, their goals, and the weights. Robots should transport each object to the goal with the same color. (e) The color bar represents the proportion of the priority of each color object at each time step.}
    \label{fig:Trajectories}
\end{figure*}

We examined the occurrence of cooperation and division in transport tasks. The trajectories of robots and objects during transport are shown in Fig.~\ref{fig:Trajectories}.

In Fig.~\ref{fig:trj-a}, each robot starts transporting objects separately. Here, Robot 0, encountering a lightweight object ($L$), can complete its transport. However, Robot 1, encountering a heavyweight object ($H$), cannot proceed with the transport. Therefore, Robot 1 abandons the transport of this object and joins Robot 2, who is transporting a medium-weight object. Next, in Fig.~\ref{fig:trj-b}, Robot 0 abandons the transport of the heavyweight object and heads towards the remaining lightweight objects. Robots 1 and 2 cooperate to advance the transport of objects. Finally, from Fig.~\ref{fig:trj-c} to Fig.~\ref{fig:trj-d}, all objects except the heavyweight ones reach their destination. Thus, the action policy by the proposed method appropriately utilizes cooperation and division according to the situation, indicating that cooperation in task allocation was achieved.

Furthermore, from Fig.~\ref{fig:trj-b} to Fig.~\ref{fig:trj-c}, the two robots were observed to transport objects in a coordinated manner by appropriately combining their forces. Compared to solitary transport, cooperative transport requires the other robot to be taken into account to adjust the force and direction applied to the object. The proposed method was confirmed to be capable of acquiring such coordination in robot control.

\subsubsection{Effectiveness of Global Communication}
We examined the effectiveness of global communication in the proposed method. From Table~\ref{tab:transported-rate}, it is evident that scenarios with communication enabled higher transport performance in all cases. Global communication primarily assists in making decisions about cooperation and division for robots and objects that are outside of local observation, suggesting that it is more effective in scenarios with a large number of robots and objects.

\section{Demonstration}
The aim of this demonstration was to verify that the policy learned in the simulation could transport the object to the target position using multiple real robots.

The positions and yaw angles of the robot and the object were observed using a motion capture system operating at \SI{120}{Hz}. The velocity and angular velocity were calculated using the measured positions and yaw angles. The control inputs were computed on a control PC and transmitted to each robot at \SI{10}{Hz} via Wi-Fi communication, using the policy learned in the simulation from these measurements.

The results are shown in Fig.~\ref{fig: real-coop} and Fig.~\ref{fig: real-indep}. Although there may be some gap between the simulation and the actual experiment, it was possible to control the object to a position close to the desired position after several trials.

\section{Discussion}
This paper introduces a multi-agent reinforcement learning framework, Task-priority Intermediated Hierarchical Distributed Policies, for coordinating multiple robots in the cooperative transport of objects of varying weights. The structured policies allow for learning control policies applicable to diverse scenarios with varying numbers of objects and robots.

In future work, our focus may shift towards refining practical robot control. This could involve delving into more complex controls, like cooperative grasping with arms capable of handling diverse shapes, for real-world application. Additionally, scalability investigation is vital for deploying these methods in large-scale logistics environments.

\bibliographystyle{IEEEtran}
\bibliography{reference}

\end{document}